\documentclass[pdflatex,sn-mathphys-num]{sn-jnl}

\usepackage{makecell}
\usepackage{graphicx}%
\usepackage{svg}
\usepackage{multirow}%
\usepackage{amsmath,amssymb,amsfonts}%
\usepackage{amsthm}%
\usepackage{mathrsfs}%
\usepackage[title]{appendix}%
\usepackage{xcolor}%
\usepackage{textcomp}%
\usepackage{manyfoot}%
\usepackage{bm}
\usepackage{booktabs}%
\usepackage[ruled,vlined]{algorithm2e}
\usepackage{listings}%
\usepackage{natbib}
\usepackage{graphicx}
\usepackage{graphicx}
\usepackage{subcaption}
\usepackage{tabularx}
\usepackage[table]{xcolor}
\usepackage[table]{xcolor} 
\usepackage{booktabs}      
\usepackage{adjustbox} 
\usepackage{multirow} 
\usepackage{tikz}
\usepackage{xcolor}
\usepackage{enumitem}
\usepackage{graphicx}
\usepackage{makecell}
\usetikzlibrary{arrows.meta, shapes, positioning}

\theoremstyle{thmstyleone}%
\theoremstyle{thmstyletwo}%

\theoremstyle{thmstylethree}%

\begin{document}

\title[Article Title]{Robustness of Graph Self-Supervised Learning to Real-World Noise: A Case Study on Text-Driven Biomedical Graphs}
\author*[1]{\fnm{Othmane} \sur{KABAL}}\email{othmane.kabal@univ-nantes.fr}

\author[1]{\fnm{Mounira} \sur{HARZALLAH}}\email{mounira.harzallah@univ-nantes.fr}

\author[1]{\fnm{Fabrice} \sur{GUILLET}}\email{Fabrice.Guillet@univ-nantes.fr}

\author[2]{\fnm{Hideaki} \sur{TAKEDA}}\email{takeda@nii.ac.jp}

\author[3]{\fnm{Ryutaro} \sur{ICHISE}}\email{ichise@iee.e.titech.ac.jp}

\affil*[1]{\orgdiv{Nantes University}, \orgname{LS2N}, \city{Nantes}, \postcode{44300}, \country{France}}

\affil[2]{\orgdiv{National Institute of Informatics}, \orgname{2-1-2 Hitotsubashi}, \orgaddress{\street{Chiyoda-ku}, \city{Tokyo} \postcode{101-8430} \country{Japan}}}
\affil[3]{\orgname{Institute of Science Tokyo}, \orgaddress{\city{Tokyo} \postcode{152-8550} \country{Japan}}}

\abstract{
Graph Self-Supervised Learning (GSSL) offers a powerful paradigm for learning graph representations without labeled data. However, existing work assumes clean, manually curated graphs. Recent advances in NLP enable the large-scale automatic extraction of knowledge graphs from text, opening new opportunities for GSSL while introducing substantial real-world noise. This type of noise remains largely unexplored, as prior robustness studies typically rely on synthetic perturbations.
To address this gap, we present the first comprehensive evaluation of GSSL methods on text-driven graphs for unsupervised term typing. We introduce Noise-Aware Text-Driven Graph GSSL (NATD-GSSL), a unified framework that combines automatic graph construction, graph refinement, and GSSL. Our evaluation follows a dual-graph protocol that contrasts a noisy graph derived from MedMentions with a clean Unified Medical Language System (UMLS) reference graph, aligned through a shared gold standard.
Our results reveal variability in robustness across both pretext tasks and Graph Neural Network (GNN) architectures. Relation reconstruction is highly sensitive to noise and benefits from well-defined schemas, whereas feature reconstruction is considerably more robust, achieving performance comparable to clean-graph settings. Contrastive objectives are generally less affected by noise but depend strongly on alignment with downstream tasks. GNN architecture also plays a critical role: bidirectional relational message-passing designs are better suited to noisy, text-driven graphs, while unidirectional relational ones perform best on clean graphs.
Overall, NATD-GSSL provides practical guidance for applying GSSL to real-world, noisy graphs and achieves up to a 7\% improvement over pretrained language model baselines. All code and benchmarks are publicly available at \url{https://github.com/OthmaneKabal/MC2GAE}.
}
\maketitle
\section{Introduction}\label{sec1}
Graph Neural Network (GNN)-based learning has witnessed rapid progress in recent years~\cite{wu2020comprehensive, dai2024comprehensive}. GNNs are now widely adopted and often preferred over purely text-based methods, as they jointly exploit textual content and the relational structure encoded in graphs \cite{10.5555/3171642.3171829, wang2024graph}. Two main paradigms have shaped this field. Supervised methods~\cite{10.1109/TKDE.2020.2981333, xiao2022graph} achieve state of the art results but require extensive labeled data, which are costly to obtain. In contrast, \textit{Graph Self-Supervised Learning} (GSSL)~\cite{liu2022graph, liu2022survey} avoids annotation by learning from pretext tasks, offering scalability and adaptability across domains.\\
A wide range of GSSL frameworks has emerged, including generative methods \cite{kipf2016variational, tan2023s2gae}, contrastive approaches~\cite{you2021graphcontrastivelearningaugmentations}, implemented using various encoder-decoder architectures~\cite{velivckovic2017graph, yang2015embeddingentitiesrelationslearning, 10.1007/978-3-319-93417-4_38}. 
However, most existing studies rely on general-purpose, well-curated graphs such as Wikidata~\cite{10.1145/2629489}, FB15k-237~\cite{toutanova2015observed}, or citation networks~\cite{sen2008collective}, which are typically considered high-quality and readily available. In practice, such resources are scarce, limiting the applicability of GSSL methods to a few domains where curated graphs exist and leaving many others unexplored. \\
To broaden the applicability of GSSL across domains, it becomes necessary to construct knowledge graphs directly from textual resources. Since manual curation is prohibitively time-consuming, automatic graph construction~\cite{zhong2023comprehensive, kabal2024enhancing} often remains the only feasible option. This process inevitably introduces noise~\cite{cai2025understanding}, posing major challenges for the robustness of GSSL methods.
While several studies have investigated the impact of noise on GNN performance~\cite{liu2022survey, Ju2024ASO, zhuang2022defending}, they primarily rely on synthetic perturbations, such as random edge deletions, additions, or adversarial attacks~\cite{zhuang2022defending}, and assess their effects on downstream performance. Nevertheless, these controlled settings rarely capture the complexity and heterogeneity of noise inherent in text-driven graphs, which tend to be structurally weak, sparse, and highly fragmented, in addition to exhibiting factual and semantic inaccuracies~\cite{hegde2015entity, mo2025kggen}.
Due to the absence of ground-truth datasets pairing raw text with large, connected knowledge graphs~\cite{cai2025understanding}, a critical issue remains largely underexplored: the sensitivity of GSSL methods to the quality of text-driven graphs.
This paper addresses this gap by systematically investigating how extraction-induced graph quality affects the downstream performance of GSSL methods.
We focus in particular on \textit{unsupervised term typing} as a downstream task a fundamental yet underexplored application of GSSL with broad implications for knowledge extraction and ontology construction~\cite{zhong2023comprehensive, kabal2024enhancing}.
We design Noise-Aware Text-Driven GSSL (NATD-GSSL), a framework that enables GSSL to operate on automatically constructed graphs by integrating a graph construction step followed by a refinement stage, which implements a set of strategies to improve graph quality.\\
To evaluate robustness in real-world scenarios, we propose a dual-graph evaluation protocol that contrasts GSSL performance on two graphs built from the same domain: a noisy graph automatically extracted from the MedMentions corpus~\cite{mohan2019medmentions}, and a clean reference knowledge graph from UMLS~\cite{bodenreider2004unified}. Because both graphs share a common set of entities and annotation schema, this setting enables a controlled and quantitative analysis of the performance gap induced by automatic graph construction. Unlike prior work based on synthetic perturbations, this evaluation reflects noise arising naturally from language ambiguity and extraction errors.
Our empirical study spans multiple GSSL Methods, implemented with various GNN architectures and trained on diverse pretext tasks in order to analyze how robustness varies across model designs and learning objectives.
The results demonstrate that relation reconstruction requires a clean graph with a well-defined schema, while feature reconstruction remains the most robust and achieves performance comparable to that of clean graphs. In contrast, contrastive approaches reveal that robustness is less dependent on graph quality and more on alignment with the downstream task.
Regarding model design, GNN architecture also plays a critical role: bidirectional relational message-passing architectures are better suited to noisy, text-driven graphs, whereas unidirectional relational architectures perform best on clean graphs.
Moreover, our refinement strategies show that graph augmentation is beneficial when the graph is sparse, while denoising even in the presence of errors may lead to degraded performance. Finally, our results also show a +7\% improvement in term typing compared to pretrained language models.\\
To summarize, our contributions are as follows:
\begin{itemize}
\item NATD-GSSL, the first framework that integrates graph construction, refinement, and GSSL into a unified process for learning from raw text.
\item A dual-graph evaluation protocol that quantitatively measures the impact of real-world noise using paired noisy and clean knowledge graphs aligned with the same gold standard.
\item A comprehensive empirical study comparing GSSL methods implemented with different GNN architectures and trained on various pretext tasks, providing practical guidance for robust model design.
\end{itemize}

The remainder of this paper is organized as follows. Section 2 reviews related work on GSSL and learning under noisy graphs. Section 3 presents the proposed NATD-GSSL framework and its modules. Section 4 details the experimental setup, Section 5 reports results and analysis, and Section 6 concludes and outlines future directions.
\section{Related Work}\label{sec2}
\subsection{Graph Self-Supervised Learning}
\label{GSSL}
Graphs provide a natural and expressive way to model relational data, where entities are represented as nodes and their interactions as edges~\cite{ZHOU202057}. 
To learn from such structures without requiring labeled data, Graph Self-Supervised Learning~\cite{liu2022graph, wang2022graph} is employed. GSSL is typically designed as an encoder-decoder framework, wherein the \textit{\textbf{encoder}} consists of stacked GNN layers that transform nodes into low-dimensional representations.
Various GNN architectures have been proposed to perform this transformation, each with distinct mechanisms for aggregating neighborhood information. GCN~\cite{kipf2016semi} performs simple neighborhood aggregation, while GAT~\cite{velivckovic2017graph} refines this through attention mechanisms, though neither considers multi-relational aspects. To address this, RGCN~\cite{10.1007/978-3-319-93417-4_38} applies relation-specific transformations but suffers from parameter explosion and unidirectional propagation. TransGCN and RotatEGCN~\cite{cai2019transgcn} address these limitations by encoding relations as translation or rotation operators, enabling bidirectional message passing with fewer parameters. Table~\ref{tab:gnn_layers} summarizes these architectures.
\begin{table}[htbp]
    \centering
    \caption{Comparison of representative GNN architectures.
    Attn: Attention mechanism; Multi-rel: multi-relational graph support;
    MPD: message passing direction ($\rightarrow$ unidirectional, $\leftrightarrow$ bidirectional).}
    \label{tab:gnn_layers}
    \setlength{\tabcolsep}{5pt}
    \renewcommand{\arraystretch}{1.15}
    \begin{tabular}{lcccl}
        \toprule
        \textbf{GNN Architecture} &
        \textbf{Attn} &
        \textbf{Multi-rel} &
        \textbf{Relation Modeling} &
        \textbf{MPD} \\
        \midrule
        GCN~\cite{kipf2016semi}
        & $\times$ & $\times$ & -- & $\rightarrow$ \\

        GAT~\cite{velivckovic2017graph}
        & \checkmark & $\times$ & -- & $\rightarrow$ \\

        RGCN~\cite{10.1007/978-3-319-93417-4_38}
        & $\times$ & \checkmark & Relation-specific & $\rightarrow$ \\

        TransGCN~\cite{cai2019transgcn}
        & Optional & \checkmark & Translation-based & $\leftrightarrow$ \\

        RotatEGCN~\cite{cai2019transgcn}
        & Optional & \checkmark & Rotation-based & $\leftrightarrow$ \\
        \bottomrule
    \end{tabular}
\end{table}\\
On the \textit{\textbf{decoder}} side, which defines the learning objective through pretext tasks, various architectures can be employed for the same task, such as standard neural networks (MLPs), GNNs, or simple scoring functions (e.g., dot product, cosine similarity). Based on the nature of these tasks, GSSL methods can be broadly divided into two main families: generative and contrastive approaches.\\
\textbf{Generative methods} 
formulate the pretext task as the reconstruction of the input graph from two complementary perspectives~\cite{liu2022graph}. First, \textit{structure reconstruction} approaches typically employ GNN-based encoders with dot-product decoders to recover the adjacency matrix~\cite{kipf2016variational, pan2019learning}.
Other approaches extend this idea to multi-relational reconstruction, aiming to recover relation types in heterogeneous graphs~\cite{10.1007/978-3-319-93417-4_38}.
Although effective for link prediction and relation extraction, these methods strongly depend on graph structure, which can limit performance in tasks where semantically similar nodes are weakly connected or linked through non-informative relations.
Second, \textit{feature reconstruction} methods focus on reconstructing node attributes~\cite{wang2017mgae, park2019symmetric}, 
which preserves semantic content when features are meaningful, but often neglects structural information. To address this limitation, third \textit{dual reconstruction} approaches jointly reconstruct both structure and features via multi-task learning with two distinct decoders~\cite{li2023multi, sun2021dual}, yielding more comprehensive embeddings nevertheless incurring a higher computational cost. These methods can be further enhanced through masking strategies~\cite{tan2023s2gae, hou2022graphmaeselfsupervisedmaskedgraph}, in which nodes or edges are partially masked and subsequently reconstructed, improving generalization and reducing overfitting.
Despite their effectiveness, generative methods tend to overfit local graph structures and often struggle to capture global contextual information, particularly in graphs with multiple disconnected components~\cite{ren2020heterogeneousdeepgraphinfomax}. Moreover, their reconstruction focused objectives often yield embeddings with limited discriminative power.\\
\textbf{Contrastive methods}, which by their nature capture global information well and produce more discriminative embeddings~\cite{10597920}, are generally developed based on the principle of mutual information maximization, where the estimated mutual information between different augmented views of the same object such as a node, a subgraph, or an entire graph is maximized \cite{zhu2020deepgraphcontrastiverepresentation, you2021graphcontrastivelearningaugmentations}. Within this family, the encoder is typically a GNN, while the decoder acts as a discriminator that estimates the level of agreement between two instances, typically using simple similarity functions such as a dot product or a bilinear function~\cite{liu2022graph}.
These methods mainly differ in their contrastive level and augmentation strategy. For instance, Graph Deep InfoMax~\cite{48921} contrasts node representations with a global graph summary, generating negative samples through node shuffling. GraphCL~\cite{you2021graphcontrastivelearningaugmentations} adopts a graph-level contrastive approach and applies a variety of data augmentations, including node dropping, edge perturbation, feature masking, and subgraph sampling. GRACE~\cite{zhu2020deepgraphcontrastiverepresentation} focuses on node-level contrastive learning by combining edge removal and feature masking to enrich local contexts, while ASP~\cite{chen2023attribute} contrasts original, attribute-based, and global views to better handle both homophilous and heterophilous graphs. While these methods produce discriminative embeddings, they remain sensitive to augmentation quality, negative sampling design, and the risk of discarding essential structural information.\\
Despite the wide range of available GSSL methods, most existing studies have evaluated them on clean, well-curated graphs. In practice, however, real-world graphs particularly those derived from text exhibit pervasive noise. The effectiveness of GSSL methods under such noisy conditions remains underexplored, especially in relation to the GNN architectures that implement these methods and the pretext tasks used to guide their learning.
\subsection{Noise in Graph Neural Networks}
\label{sec:GNN_noisy}
The quality of the input graph plays a crucial role in the effectiveness of GNNs ~\cite{Ju2024ASO}. In practice, graphs are rarely perfect and often suffer from various types of noise, which are commonly categorized into two main types~\cite{paulheim2016knowledge, liu2022survey}. 
\textit{Structural noise} refers to inconsistencies in the graph topology, such as missing or spurious edges, that distort the genuine relationships between nodes ~\cite{rong2019dropedge, yuan2023self}. Missing edges increase graph sparsity and hinder effective information propagation, while spurious edges may introduce misleading connections, leading to over-smoothing and incorrect message aggregation. Together, these issues disrupt the message-passing mechanisms of GNNs and negatively impact model performance~\cite{fox2019robust}. \textit{Node-level} noise arises from erroneous, missing, or incomplete node attributes~\cite{Ju2024ASO}. Such noise reduces the informativeness of node features and impairs neighborhood aggregation, ultimately limiting a model’s ability to learn accurate and meaningful node representations~\cite{liu2022survey}.\\
To study the impact of noise on GNNs, a common methodology in the literature consists in introducing controlled synthetic perturbations into the input graph \cite{wang2023user, ennadir2024simple}. These perturbations typically include randomly adding or removing edges, corrupting node features (e.g., by injecting Gaussian noise or masking attributes), or applying more sophisticated adversarial attacks \cite{zhuang2022defending, guo2022learning}. Model robustness is then assessed on downstream tasks, where a noticeable degradation in performance is consistently observed, highlighting the high sensitivity of GNNs to graph corruption even under moderate perturbations \cite{jin2020graph}.
While these approaches are well suited for controlled experimental settings, they are largely conducted on mono-relational benchmark graphs such as Cora, CiteSeer, and PubMed \cite{sen2008collective}, which are relatively simple and carefully curated. Alternatively, some studies~\cite{wang2019robust, dong-etal-2025-refining} rely on existing multi-relational graphs that are manually or semi-automatically constructed from structured sources, such as DBpedia and Freebase \cite{auer2007dbpedia, bollacker2008freebase}. In both cases, these methodologies assume access to a clean, high-quality underlying graph, which limits their applicability to real-world scenarios where the graph is inherently noisy, as is typically the case for graphs automatically constructed from textual data.
In contrast to curated graphs where noise is relatively simple and more commonly stems from fact obsolescence, human editing errors, or source alignment issues rarely involving severe violations of semantic constraints due to the limited expressiveness of the underlying ontologies~\cite{paulheim2016knowledge}. Noise in text-driven graphs predominantly originates from upstream NLP pipelines\cite{zhong2023comprehensive}, including errors in entity recognition, entity disambiguation, relation extraction, and coreference resolution. These errors often lead to inconsistent triples (e.g., \textit{Barack Obama}, \textit{siblingOf}, \textit{White House}), factually false triples (e.g., \textit{Boston}, \textit{capitalOf}, \textit{USA}), or overly generic triples (e.g., \textit{family}, \textit{residesIn}, \textit{New York})~\cite{mihindukulasooriya2017towards}, which are characteristic failure modes of automatic extraction systems. 
Moreover, text-driven graphs are particularly affected by entity duplication, arising from spelling variations, unresolved aliases, or incomplete entity normalization, which  leads to the creation of artificial nodes an issue that is largely absent from curated knowledge graphs~\cite{cai2025understanding} and further results in increased sparsity and severe graph fragmentation~\cite{mo2025kggen}.
Finally, whereas curated graphs typically prioritize precision over coverage, automatically constructed graphs often favor high recall, leading to substantially higher levels of structural and semantic noise~\cite{faralli2023benchmark, kabal2024enhancing}.

Given these characteristics, existing NLP research lacks ground-truth datasets pairing raw text with large, connected knowledge graphs, limiting systematic analysis of how realistic extraction noise propagates to downstream graph learning tasks~\cite{cai2025understanding}. 
As a result, the impact of extraction-induced noise on GNNs particularly GSSL methods remains largely unexplored.
To address this gap, our study addresses the above-mentioned limitations by
(i) integrating the processes of knowledge graph construction and graph self-supervised learning into a unified framework NATD-GSSL, enabling systematic analysis of noise propagation; and
(ii) introducing a novel dual-graph evaluation protocol that contrasts noisy, text-driven graphs with clean, existing graphs in order to quantitatively assess the performance degradation induced by the graph construction process.
\section{NATD-GSSL Framework}
\begin{figure}[ht]
  \centering
  \includegraphics[width=\linewidth]{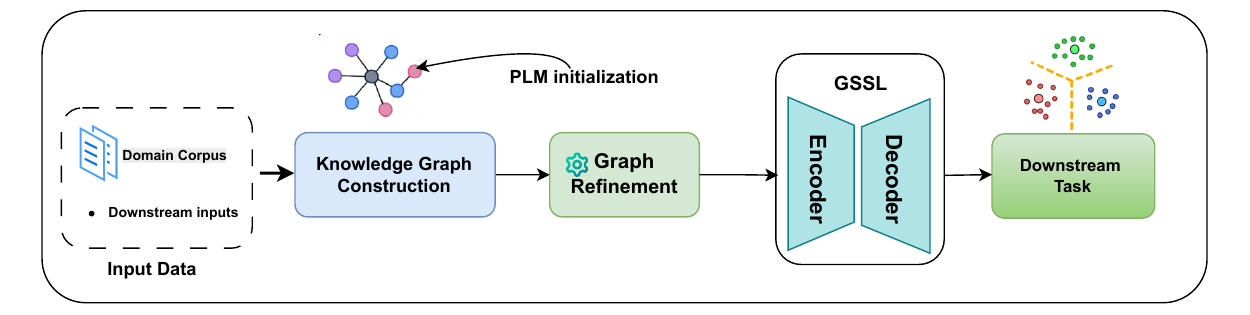}
  \caption{Overview of the NATD-GSSL Framework.}
  \label{fig:archi_pipeline}
\end{figure}

This section presents the NATD-GSSL framework, which extends the applicability of GSSL methods to settings where only raw textual data is available. Unlike conventional approaches that assume access to pre-existing, clean graphs, NATD-GSSL explicitly integrates knowledge graph construction into the learning pipeline, thereby enabling systematic analysis of extraction-induced noise and its impact on downstream tasks. Furthermore, the framework incorporates a dedicated component that mitigates this noise through a set of graph refinement strategies. As illustrated in Figure~\ref{fig:archi_pipeline}, the NATD-GSSL framework consists of four main components. 

\subsection{Input Data and Knowledge graph Construction}
NATD-GSSL takes as input a domain-specific corpus consisting of a collection of unstructured textual documents relevant to the target domain in which GSSL is to be applied. The framework accepts additional inputs specific to the downstream task, which may include entities, concepts, or labels, along with any auxiliary structures required by the task formulation.
Based on these inputs, a \emph{knowledge graph (KG)} is automatically constructed using the general-purpose method General
Text To Knowledge Graph (\textsc{GT2KG})~\cite{kabal2024enhancing}, which ensures the domain independence of the framework. Formally, the KG is defined as
\[
G = (V, E, X, R),
\]
where $V$ denotes the set of nodes and $E \subseteq V \times R \times V$ represents the set of relational edges. The node feature matrix $X \in \mathbb{R}^{|V| \times d}$ is obtained by encoding node-associated textual information using pretrained language models (PLMs). The set $R$ denotes the collection of relation types. Each edge is represented as a triplet $(v_i, r, v_j)$, where $v_i, v_j \in V$ are the source and target nodes, respectively, and $r \in R$ specifies the semantic relation connecting them.

\subsection{Graph Refinement}
\label{sec:graph_refine}
The constructed graph, as discussed in Section~\ref{sec:GNN_noisy}, suffers from several imperfections. These include structural issues related to fragmentation and incompleteness, as well as errors in the semantic content.\\
The graph refinement component proposes a set of strategies aimed at improving the overall quality of the graph. These strategies are grouped into two main categories: \textit{enrichment} and \textit{cleaning}.
\paragraph{Enrichment}
Enhances the graph by adding new nodes and/or edges in order to mitigate the structural weaknesses and reinforce connectivity.
For this purpose, we propose: \\
\textbf{Rule-based \textit{is-a} augmentation} 
introduces new nodes and is-a relations by exploiting morphosyntactic patterns in multi-word terms to infer missing hierarchical relations.
This augmentation
reduces graph fragmentation and improves message passing, while minimizing the risk of introducing semantic errors.
We distinguish two types of rules depending on whether the term contains the preposition "of". The corresponding rules and illustrative examples are presented in Table~\ref{tab:augmentation_rules}.
\begin{table}[ht]
\centering
\caption{Rules for is-a Augmentation from Composed Terms}
\label{tab:augmentation_rules}
\small
\begin{tabular}{|l|l|l|}
\hline
\textbf{Rule Type} & \textbf{Syntactic Structure} & \textbf{\begin{tabular}[t]{@{}l@{}}Examples with\\Generated Relationships\end{tabular}} \\
\hline
\textbf{Without "of"} & 
\begin{tabular}[t]{@{}l@{}}
Term: A B C \\
Root: C (last word) \\
Rules: \\
• (B C, is-a, C) \\
• (A B C, is-a, B C)
\end{tabular} & 
\begin{tabular}[t]{@{}l@{}}
Term: Stem cell therapy \\
Relationships: \\
• (Cell therapy, is-a, therapy) \\
• (Stem cell therapy, is-a,\\
\phantom{•} cell therapy)
\end{tabular} \\
\hline
\textbf{With "of"} & 
\begin{tabular}[t]{@{}l@{}}
Term: C of B of A \\
Root: C (first word before "of") \\
Rules: \\
• (B of C, is-a, C) \\
• (C of B of A, is-a, B of C)
\end{tabular} & 
\begin{tabular}[t]{@{}l@{}}
Term: Disease of cell physiology \\
Relationships: \\
• (Disease of cell physiology,\\
\phantom{•} is-a, disease)
\end{tabular} \\
\hline
\end{tabular}
\end{table}



\paragraph{Cleaning}
This step aims to remove noise from the graph content, affecting both relations and nodes. To this end, we adopt a removal-based refinement strategy proposed by~\cite{dong-etal-2025-refining}. This method relies on large language models (LLMs) and has been validated by its original authors, ensuring the reliability of the underlying validation process. We follow this established framework without modifying its core decision mechanism.
The approach operates in two stages: noise detection and filtering. Given a triple $(E_1, R, E_2)$ extracted from a source sentence $s$, an LLM is employed to assess whether the relation expressed by the triple is supported by the sentence context and the model’s linguistic knowledge. Formally, this validation step is modeled as a binary decision function:
\[
\mathcal{V}_{\text{LLM}}(E_1, R, E_2 \mid s) \rightarrow \{0,1\},
\]
where a value of $1$ indicates that the relation is semantically supported by $s$, while $0$ denotes a noisy or unsupported triple.\\
Based on this decision, triples identified as noisy are removed from the graph, while only validated triples are retained.


\subsection{GSSL and Downstream Task}
Once the graph is refined, the text-driven graph is passed to the GSSL component, which aims to learn node embeddings by leveraging both the structural and semantic information encoded in the graph. To this end, a graph encoder \( f_\theta \) is employed, producing:
\[
\mathbf{H} = f_\theta(G) = f_\theta(\mathbf{X}, \mathcal{R}, \mathcal{E}_R),
\]
where $\mathbf{H} \in \mathbb{R}^{|V| \times d}$ denotes the set of node representations learned in a self-supervised manner through a pretext task using a decoder $f_\phi$, without requiring any labeled data.
Finally the learned representations $\mathbf{H}$ are subsequently exploited by downstream tasks through task-specific inference functions. Depending on the application, these embeddings can be used for various tasks such as node classification, clustering, or link prediction, either directly or via lightweight task-specific heads. 
\section{Experiments}
In this section, we present our experimental study that instantiates the NATD-GSSL framework to investigate the impact of graph construction quality on the performance of Graph GSSL methods. Unless otherwise stated, all experimental settings correspond to specific instantiations of NATD-GSSL, differing only in the graph refinement stage. We first introduce the downstream task adopted in this study, followed by the proposed dual-graph evaluation protocol and a description of the datasets and graph variants used in our experiments. We then detail the different GSSL methods considered in this study, as well as the experimental setting.
\subsection{Downstream Task: Term Typing}
We adopt \emph{term typing} as the downstream task, which consists in assigning each target term to one or more predefined semantic types. This task is particularly relevant in graph-based settings, as it directly evaluates the semantic coherence of learned node representations and is widely used in knowledge graph construction and enrichment.
For this task, NATD-GSSL requires two additional inputs:
(i) a set of \emph{target terms} to be typed, which appear in the corpus and are therefore contextualized by it; and
(ii) a predefined set of \emph{semantic types} to which these target terms are to be assigned.
Once the graph is constructed to explicitly include both target terms and predefined semantic types as nodes, node representations are learned through the GSSL component of NATD-GSSL. The resulting embeddings $\mathbf{H}$ are then used to perform type assignment. 
We adopt a \emph{nearest type assignment} strategy based on embedding similarity.
Given the embeddings produced by the GSSL encoder, 
each target node $v_i$ is assigned the semantic type $t_j \in \mathcal{T}$ whose embedding is most similar to that of $v_i$, measured using cosine similarity:
\begin{equation}
\phi(v_i) = \arg\max_{t_j \in \mathcal{T}} \cos(\mathbf{H}_{v_i}, \mathbf{H}_{t_j}),
\label{eq:phi}
\end{equation}
where $\mathbf{H}_{v_i}$ and $\mathbf{H}_{t_j}$ denote the embeddings of the target term node and the semantic type node, respectively.

\subsection{Dual-Graph Evaluation Protocol}
\label{subsec:dual_graph_eval}
\begin{figure}[ht]
  \centering
  \includegraphics[width=\linewidth]{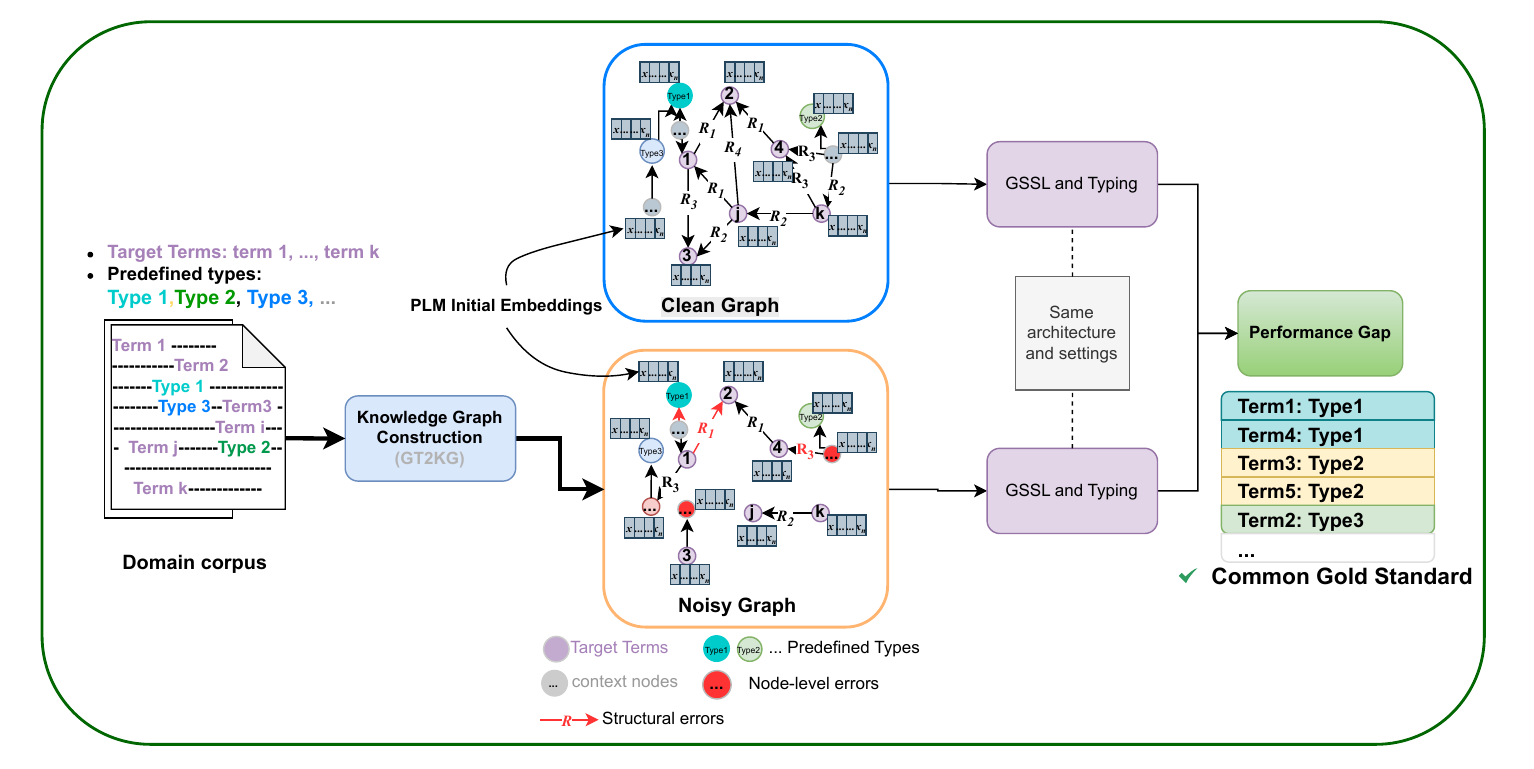}
  \caption{Dual-graph evaluation protocol}
  \label{fig:dual_eval}
\end{figure}

To quantify the impact of graph quality on downstream performance, we introduce a \textit{\textbf{dual-graph evaluation protocol}} based on two graphs derived from the same domain.
The first graph is automatically constructed from raw text, and as such, it contains various forms of real-world noise including structural fragmentation, spurious or missing edges, and noisy or incomplete node attributes. This graph includes a set of target terms to be typed, along with predefined semantic type nodes, following the NATD-GSSL framework.\\
The second graph is a high-quality reference graph, curated by domain experts. It includes the same set of target terms and semantic type nodes as the noisy graph, which enables a direct and meaningful comparison.\\
All models are trained and evaluated using the same experimental setup including architecture, dimensionality, optimization settings, and evaluation procedure. The only difference between runs is the input graph.
By applying GSSL independently to both graphs and evaluating the same downstream task (unsupervised term typing), we can measure the performance gap that is, the loss in performance attributable to the noise present in the automatically constructed graph.
Figure~\ref{fig:dual_eval} illustrates this dual-graph evaluation protocol.

\subsection{Data and Graph Variants}
\label{subsec:data_graphs}
To support the dual-graph evaluation protocol, this section describes the data sources and the different graph variants used in our experiments.
\paragraph{Input Corpus}
We use the \textbf{MedMentions corpus}~\cite{mohan2019medmentions}, a large-scale biomedical dataset originally designed for concept recognition and entity typing. This corpus is particularly well suited to our experimental setting for two main reasons.
First, it provides a large collection of domain-specific terms requiring semantic typing, which naturally supports the unsupervised term typing task considered in this work.
Second, MedMentions is annotated according to the \textbf{UMLS semantic type system}, enabling direct alignment with the UMLS knowledge graph.

\paragraph{Graph Variants.}
Based on this corpus, we construct and evaluate the following graph variants:

\begin{itemize}
    \item \textbf{Clean Reference Graph.} A high-quality biomedical knowledge graph validated by domain experts derived from the UMLS-NCI Thesaurus~\cite{bodenreider2004unified}. It is characterized by a well-defined schema, dense connectivity, absence of fragmentation, and coherent semantic organization. This graph serves as the reference graph in our dual-graph evaluation protocol.
    
    \item \textbf{Noisy Graph.} Automatically constructed from the MedMentions corpus using the GT2KG approach. This graph exhibits severe structural weaknesses, including extreme fragmentation, low average node degree, and sparse connectivity.
    Examples of noisy triplets in the constructed graph are illustrated in Table~\ref{tab:noise_examples}.
    \begin{table}[htbp]
    \centering
    \small
    \setlength\extrarowheight{2pt}
    \caption{Examples of real-world noise in text-driven graph}
    \label{tab:noise_examples}
    \begin{tabularx}{\textwidth}{|p{2.4cm}|X|p{2.5cm}|X|}
    \hline
    \textbf{Noise Type} &
    \textbf{Extracted Triplet (or Mis sing Element)} &
    \textbf{Noise Category} &
    \textbf{Explanation} \\ \hline
    
        Erroneous Link &
    (Plant, \textit{related\_to}, Intellectual Product) &
    Structural &
    Two nodes that should not be connected \\ \hline
        
    Missing Node &
    \textit{Missing node:} "Cancer Genetics Network" present in text but not extracted &
    Node-level &
    Important concept missing, degrading subgraph completeness. \\ \hline
    
    Generic Entity &
    (study, \textit{associated\_with}, outcome) &
    Node-level &
    Overly frequent generic nodes form hubs, biasing message passing. \\ \hline
    
    Boundary Error &
    (avian influenza and Ebola virus hemorrhagic fever, \textit{cause}, symptoms) &
    Node-level &
    Two entities are concatenated due to conjunction misinterpretation. \\ \hline
    Entity Ambiguity &
    (conventional lipid extraction technique, \textit{compare}, traditional method of lipid extraction) &
    Semantic &
    Two expressions refer to the same concept but appear as different nodes. \\ \hline
    Wrong Relation &
    (heavy metal, \textit{is}, abiotic stress) &
    Structural \& Semantic &
    Heavy metals cause stress but are not a subtype of abiotic stress. \\ \hline
    
    Node-level \& Structural \& Semantic  &
    (mesenchymal stem cell MTH-68H, \textit{provide}, Newcastle disease virus novel therapeutic approach) &
    Node-level \& Semantic \& Structural &
    Object mixes true concept (virus) with irrelevant addon “novel therapeutic approach”. \\ \hline
    
    Wrong Relation \& Wrong Entity &
    (median age, \textit{is}, 63 year) &
    Node-level \& Structural \& Semantic &
    “is-a’’ is incorrect; numeric values are not valid entities. \\ \hline
    \end{tabularx}
\end{table}

    \item \textbf{Enriched Graph.} Obtained by applying enrichment-based refinement strategies to the noisy graph, aimed at alleviating structural deficiencies and reinforcing connectivity through the insertion of additional hierarchical relations.
    
    \item \textbf{Cleaned Graph.} Generated by applying the cleaning refinement strategy, in which detected noisy triples are filtered out.
    We follow the same prompting strategy protocol as in \cite{dong-etal-2025-refining}, using \textit{DeepSeek} with reasoning enabled.
    \item \textbf{Combined Refined Graph.} Generated by applying both refinement strategies: first enrichment, followed by cleaning, in order to also check the added triplets.
\end{itemize}

Key topological statistics of all graph variants are reported in Table~\ref{tab:graph_topology_stats}.

\begin{table}[htbp]
    \centering
    \caption{Graph Statistics with Topological Measures.\\
    $n_{\text{comp}}$ → number of connected components, 
    $r_{\text{giant}}$ → giant component ratio, 
    avg\_deg → average node degree.
    }
    \label{tab:graph_topology_stats}
    \begin{tabular}{|l|r|r|r|r|r|r|}
    \hline
    \textbf{Graph Variant} & \textbf{Nodes} & \textbf{Edges} & \textbf{Relations} & 
    $\boldsymbol{n_{\text{comp}}}$ & $\boldsymbol{r_{\text{giant}}}$ & 
    \textbf{avg\_deg} \\
    \hline
    
    \makecell[l]{\textbf{Noisy Graph}} 
    & 37,142 & 34,322 & 76 & 6,987 & 0.51 & 1.85 \\
    \hline
    
    \makecell[l]{\textbf{Clean Reference }\\\textbf{Graph}} 
    & 184,574 & 1,258,931 & 110 & 1 & 1.00 & 13.60 \\
    \hline
    
    \makecell[l]{\textbf{Enriched Graph}} 
    & 58,483 & 80,193 & 76 & 989 & 0.94 & 2.74 \\
    \hline
    
    \makecell[l]{\textbf{Cleaned Graph}} 
    & 30,007 & 25,932 & 76 & 6,532 & 0.43 & 1.72 \\
    \hline
    
    \makecell[l]{\textbf{Combined Refined}\\\textbf{Graph}} 
    & 50,895 & 60,984 & 76 & 3,210 & 0.82 & 2.39 \\
    \hline
    \end{tabular}
\end{table}

\paragraph{Gold Standard and Evaluation Scope.}
All graph variants share an identical set of target terms, forming a gold standard composed of \textbf{1,040 nodes} equally distributed across \textbf{eight semantic types}. Evaluation is conducted exclusively on this shared set of target terms.\\
Performance is evaluated using \textbf{Accuracy}, \textbf{macro-Precision}, and \textbf{macro-F1-score}. Recall is not reported separately, as it is equivalent to Accuracy under the balanced class distribution of the gold standard.

\subsection{GSSL Methods and Baseline} 
We systematically explored a broad range of GSSL methods, covering both generative and contrastive paradigms. These methods are trained on different pretext tasks, including relation reconstruction, feature reconstruction, and contrastive learning, each implemented using various encoder-decoder architectures built upon different layers, such as GNNs and MLPs.
This systematic exploration allows us to assess how architectural choices influence robustness to real-world graph noise. In total, 36 distinct GSSL configurations were evaluated. The complete set of explored variants is summarized in Table~\ref{tab:GSSL_variants}.
\begin{table}[!h]
    \centering
    \caption{GSSL methods considered. Feat. Rec. → Feature Reconstruction, Rel. Rec. → Relation Reconstruction}
    \label{tab:GSSL_variants}
    \small
    \begin{tabular}{|l|l|l|l|l|}
    \hline
    \textbf{Pretext task} & \textbf{Category} & \textbf{Encoders} & \textbf{Decoders} & \textbf{Loss} \\
    \hline
    Feat. Rec. & Generative & 
    \begin{tabular}[t]{@{}l@{}}GCN~\cite{kipf2016semi}, GAT~\cite{velivckovic2017graph}\\RGCN~\cite{10.1007/978-3-319-93417-4_38}, TransGCN/\\RotatEGCN~\cite{cai2019transgcn}\end{tabular} & 
    \begin{tabular}[t]{@{}l@{}}GCN~\cite{kipf2016semi}, GAT~\cite{velivckovic2017graph}, MLP\\RGCN~\cite{10.1007/978-3-319-93417-4_38}, TransGCN/\\RotatEGCN~\cite{cai2019transgcn}\end{tabular} & 
    MSE \\
    \hline
    Rel. Rec. & Generative & RGCN~\cite{10.1007/978-3-319-93417-4_38} & DistMult~\cite{yang2015embeddingentitiesrelationslearning} & MCE \\
    \hline
    Contrastive & Contrastive & 
    \begin{tabular}[t]{@{}l@{}}GCN~\cite{kipf2016semi}, GAT~\cite{velivckovic2017graph}\\RGCN~\cite{10.1007/978-3-319-93417-4_38}, TransGCN/\\RotatEGCN~\cite{cai2019transgcn}\end{tabular} & 
    \begin{tabular}[t]{@{}l@{}}contrastive scoring~\cite{zhu2020deepgraphcontrastiverepresentation} \end{tabular} & 
    InfoNCE~\cite{zhu2020deepgraphcontrastiverepresentation} \\
    \hline
    \end{tabular}
\end{table}

To establish a meaningful point of comparison, we define a non-graph baseline referred to as the Textual PLM Baseline (PLM-Only). This baseline relies solely on pretrained textual semantics without using any graph structure. Each term is represented using static embeddings derived from a pretrained language model, and semantic types are assigned based on cosine similarity in the embedding space using Eq~\ref{eq:phi}. This baseline serves two key purposes: first, it provides a lower bound that reflects the performance achievable without leveraging relational information; second, it allows us to isolate the contribution of GSSL, as the same pretrained embeddings are used to initialize the node representations in all graph-based configurations.

\subsection{Experimental Setup}
All experiments were conducted on an NVIDIA RTX 4500 Ada GPU (24 GB VRAM) with CUDA 12.6, using PyTorch and PyTorch Geometric as the primary frameworks. \\
Node and relation features were initialized using pre-trained language models, with \textit{sentence-transformers/all-MiniLM-L6-v2} selected after comparative evaluation across general-purpose, biomedical, and semantic similarity PLM families (details in Appendix~\ref{app:plm_init}).\\
Training was performed using mini-batch mode with GraphSAGE~\cite{hamilton2017inductive} neighborhood sampling, retaining 200 neighbors per layer.  All models were trained for 50 epochs and repeated three times under different random seeds to ensure reproducibility.
The encoder architecture consisted of two hidden layers with dimensional configurations of [384, 256] and [256, 128], while batch sizes were set to 256 or 512 depending on the graph size.
For RGCN layers, the number of basis matrices (\texttt{num\_bases}) was set between 5 and 10 to balance expressiveness and computational efficiency. 
\section{Results and Analysis}
In this section, we present the results of the conducted experiments, followed by an in-depth analysis. 
This analysis aims to answer the following research questions:

\begin{itemize}
    \item[\textbf{Q1.}] How does the performance of GSSL on a text-driven graph compare to its performance on a clean graph?
    \item[\textbf{Q2.}] 
    Is the relative performance of GSSL method (pretext task, encoder-decoder architecture) influenced by graph quality?
    \item[\textbf{Q3.}] What is the contribution of graph refinement methods, particularly enrichment versus cleaning techniques?
\end{itemize}
\subsection{Results Overview}
The results obtained with the different GSSL methods on all graph variants are presented in Table~\ref{tab:gssl_results}, where the best-performing models for each pretext task are highlighted. For feature reconstruction, however, the top three configurations are reported due to the limited performance variation observed across models.\\
\newcommand{\mstd}[2]{\ensuremath{#1_{\scriptsize\pm\,#2}}}
\newcommand{\mstdB}[2]{\ensuremath{\bm{#1}_{\scriptsize\pm\,#2}}}

\begin{figure}[htbp]
    \centering
    \hspace*{\fill}
    \begin{subfigure}[t]{0.32\textwidth}
        \centering
        \includegraphics[width=\textwidth]{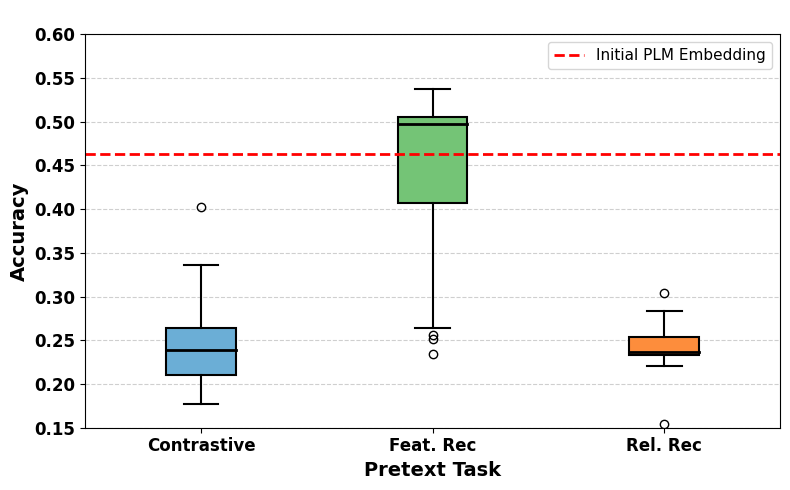}
        \caption{Noisy Graph}
        \label{fig:noisy}
    \end{subfigure}
    \hfill
    \begin{subfigure}[t]{0.32\textwidth}
        \centering
        \includegraphics[width=\textwidth]{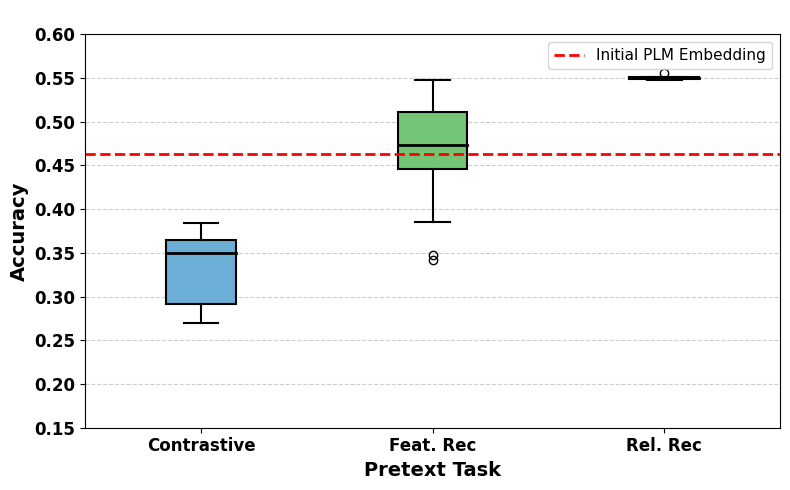}
        \caption{Clean Reference Graph}
        \label{fig:clean_ref}
    \end{subfigure}
    \hspace*{\fill}
    
    \vspace{0.3cm}
    
    \begin{subfigure}[t]{0.32\textwidth}
        \centering
        \includegraphics[width=\textwidth]{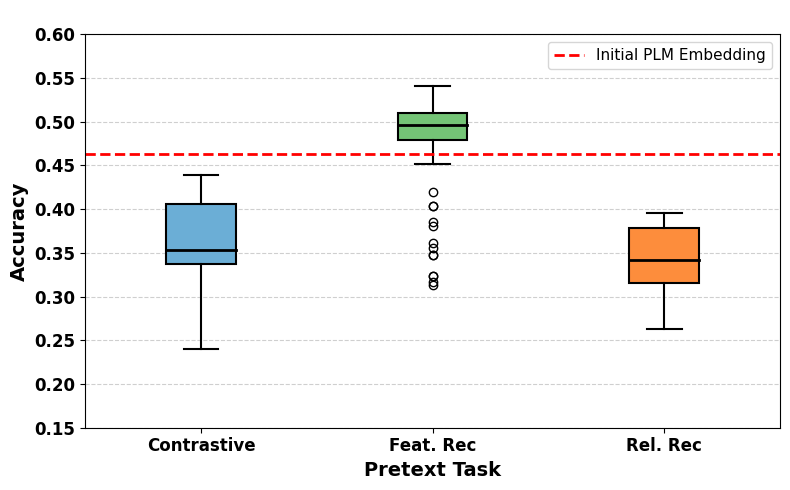}
        \caption{Enriched Graph}
        \label{fig:enriched}
    \end{subfigure}
    \hfill
    \begin{subfigure}[t]{0.32\textwidth}
        \centering
        \includegraphics[width=\textwidth]{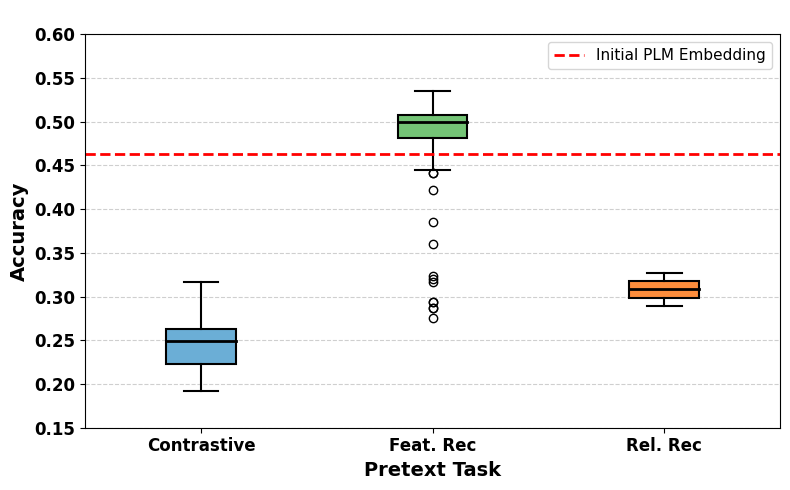}
        \caption{Cleaned Graph}
        \label{fig:cleaned}
    \end{subfigure}
    \hfill
    \begin{subfigure}[t]{0.32\textwidth}
        \centering
        \includegraphics[width=\textwidth]{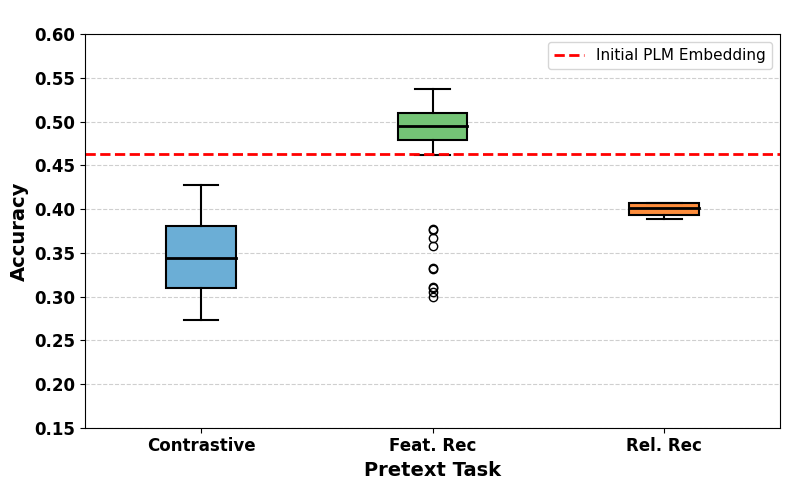}
        \caption{Combined Refined Graph}
        \label{fig:combined}
    \end{subfigure}
    
    \caption{Accuracy distribution across pretext tasks under different data settings. The red dashed line indicates the performance of the initial PLM embedding.}
    \label{fig:pretext_accuracy_all}
\end{figure}

\newcommand{\attn}{\ensuremath{_{\scriptsize\text{attn}}}}
\newcommand{\conv}{\ensuremath{_{\scriptsize\text{conv}}}}
\begin{table*}[htbp]
\centering
\caption{Performance of GSSL model variants per graph type. Feat. Rec. $\rightarrow$ Feature Reconstruction, Rel. Rec. $\rightarrow$ Relation Reconstruction, Acc. $\rightarrow$ Accuracy, Prec. $\rightarrow$ Precision. Best results per graph variant in \textbf{bold}.}
\label{tab:gssl_results}
\small
\begin{tabular}{|l|l|l|c|c|c|}
\hline
\textbf{Pretext Task} & \textbf{Encoder} & \textbf{Decoder} & \textbf{Acc.} & \textbf{F1} & \textbf{Prec.} \\
\hline\hline
\multicolumn{6}{|c|}{\textit{Baseline (PLM initial embedding)}} \\
\hline
* & all-MiniLM-L6-v2 & * & 0.4634 & 0.4371 & 0.4730 \\
\hline\hline
\multicolumn{6}{|c|}{\textbf{Clean Reference Graph}} \\
\hline
Feat. Rec. & RGCN & MLP &
\mstd{0.5480}{0.008} & \mstd{0.5180}{0.010} & \mstd{0.5345}{0.009} \\
Feat. Rec. & TransEGCN\conv & GAT &
\mstd{0.5442}{0.009} & \mstd{0.5254}{0.011} & \mstd{0.5342}{0.010} \\
Feat. Rec. & RotatEGCN\conv & MLP &
\mstd{0.5413}{0.007} & \mstdB{0.5287}{0.009} & \mstdB{0.5400}{0.008} \\
Rel. Rec. & RGCN & DistMult &
\mstdB{0.5577}{0.014} & \mstd{0.5234}{0.016} & \mstd{0.5290}{0.015} \\
Contrastive & GAT & Contrastive scoring &
\mstd{0.3836}{0.028} & \mstd{0.3629}{0.031} & \mstd{0.3688}{0.029} \\
\hline\hline
\multicolumn{6}{|c|}{\textbf{Noisy Graph}} \\
\hline
Feat. Rec. & TransEGCN\conv & RotatEGCN\attn &
\mstdB{0.5355}{0.014} & \mstdB{0.5197}{0.017} & \mstd{0.5275}{0.015} \\
Feat. Rec. & TransEGCN\attn & TransEGCN\attn &
\mstd{0.5346}{0.015} & \mstd{0.5144}{0.018} & \mstd{0.5125}{0.016} \\
Feat. Rec. & TransEGCN\conv & TransEGCN\attn &
\mstd{0.5336}{0.013} & \mstd{0.5148}{0.017} & \mstdB{0.5402}{0.014} \\
Rel. Rec. & RGCN & DistMult &
\mstd{0.3038}{0.026} & \mstd{0.2776}{0.030} & \mstd{0.2802}{0.028} \\
Contrastive & RotatEGCN\conv & Contrastive scoring &
\mstd{0.4028}{0.041} & \mstd{0.3792}{0.045} & \mstd{0.4117}{0.039} \\
\hline\hline
\multicolumn{6}{|c|}{\textbf{Enriched Graph}} \\
\hline
Feat. Rec. & RotatEGCN\conv & MLP &
\mstdB{0.5413}{0.009} & \mstdB{0.5223}{0.012} & \mstdB{0.5490}{0.010} \\
Feat. Rec. & RGCN & MLP &
\mstd{0.5394}{0.010} & \mstd{0.5162}{0.013} & \mstd{0.5349}{0.011} \\
Feat. Rec. & TransEGCN\conv & RGCN &
\mstd{0.5288}{0.012} & \mstd{0.4965}{0.015} & \mstd{0.5176}{0.013} \\
Rel. Rec. & RGCN & DistMult &
\mstd{0.3952}{0.020} & \mstd{0.3384}{0.024} & \mstd{0.4265}{0.022} \\
Contrastive & TransEGCN\conv & Contrastive scoring &
\mstd{0.4394}{0.034} & \mstd{0.4316}{0.037} & \mstd{0.4699}{0.035} \\
\hline\hline
\multicolumn{6}{|c|}{\textbf{Cleaned Graph}} \\
\hline
Feat. Rec. & TransEGCN\attn & TransEGCN\attn &
\mstdB{0.5346}{0.011} & \mstdB{0.5189}{0.014} & \mstdB{0.5434}{0.012} \\
Feat. Rec. & TransEGCN\conv & TransEGCN\attn &
\mstd{0.5307}{0.012} & \mstd{0.5163}{0.015} & \mstd{0.5400}{0.013} \\
Feat. Rec. & RotatEGCN\conv & TransEGCN\conv &
\mstd{0.5307}{0.011} & \mstd{0.5128}{0.014} & \mstd{0.5251}{0.012} \\
Rel. Rec. & RGCN & DistMult &
\mstd{0.3952}{0.021} & \mstd{0.3384}{0.025} & \mstd{0.4265}{0.023} \\
Contrastive & RotatEGCN\conv & Contrastive scoring &
\mstd{0.3269}{0.029} & \mstd{0.3125}{0.032} & \mstd{0.3292}{0.030} \\
\hline\hline
\multicolumn{6}{|c|}{\textbf{Combined Refined Graph}} \\
\hline
Feat. Rec. & TransEGCN\attn & TransEGCN\conv &
\mstdB{0.5375}{0.012} & \mstd{0.5186}{0.015} & \mstd{0.5180}{0.014} \\
Feat. Rec. & TransEGCN\conv & GAT &
\mstd{0.5355}{0.013} & \mstdB{0.5224}{0.016} & \mstdB{0.5445}{0.014} \\
Feat. Rec. & RGCN & MLP &
\mstd{0.5042}{0.015} & \mstd{0.5042}{0.017} & \mstd{0.5183}{0.016} \\
Rel. Rec. & RGCN & DistMult &
\mstd{0.4067}{0.019} & \mstd{0.3497}{0.023} & \mstd{0.3955}{0.021} \\
Contrastive & RotatEGCN\conv & Contrastive scoring &
\mstd{0.4278}{0.033} & \mstd{0.4086}{0.036} & \mstd{0.4165}{0.034} \\
\hline
\end{tabular}
\end{table*}
The results show a noticeable variation in performance depending on both the encoder-decoder architecture and the pretext task. While some configurations lead to significant improvements up to +7\% in Accuracy compared to PLM\_Init, others exhibit a clear degradation down to -16\%, with accuracy gaps reaching up to 0.25 between the best and worst models under the same typing task. 
Since Table~\ref{tab:gssl_results} only reports the best-performing models per task, Figure~\ref{fig:pretext_accuracy_all} complements it by illustrating the full range of results obtained across all encoder-decoder combinations for each pretext task. This variation highlights the strong influence of two factors: the choice of pretext task itself, and, within the same task, the specific architectural design of the encoder-decoder components. We therefore conduct a two-fold analysis based on these axes of variation.
\subsection{Pretext task Analysis}
\begin{itemize}
    \item \textbf{Feature Reconstruction:}
The feature reconstruction pretext task demonstrates the highest robustness across all graph variants. It consistently yields strong performance, achieving an approximate 8\% improvement in accuracy compared to PLM-initialized embeddings, with a minimal performance gap of less than 1\% between noisy and clean reference graph settings. This stability suggests that the task generalizes well to varying noise levels, as it aims to reconstruct the semantic node features originally derived from PLMs while incorporating contextual information through neighborhood aggregation. Even in the presence of noisy neighborhoods such as weak connections or erroneous edges the pretext task seeks to preserve the initial embeddings. 
As illustrated in Figure~\ref{fig:pretext_accuracy_all}, however, performance stability varies across graph qualities: results on the Noisy Graph exhibit higher variance across architectural choices, indicating that feature reconstruction is more sensitive to model design in structurally deficient settings. In contrast, the Clean Reference Graph leads to more consistent results across architectures, highlighting the stabilizing effect of well-structured relational data. Moreover, the refined graphs (enriched, cleaned, and combined) demonstrate both greater stability and overall performance gains over the PLM baseline, further confirming that even partial graph refinement can mitigate architectural sensitivity and reinforce the effectiveness of feature reconstruction.
However, this strategy presents a double-edged sword: while it ensures stability, it may perform poorly when the initial embeddings themselves are suboptimal.
As shown in Figure~\ref{fig:transition_matrices}, correct classifications issues from  initial embeddings are largely preserved in the final predictions (65.7\% on the clean reference graph and 86.3\% on the noisy graph), whereas incorrect initial classifications are rarely corrected (44.8\% and 25.3\%, respectively). 
\begin{figure}[htbp]
    \centering
    \captionsetup{font=small}

    \begin{subfigure}[b]{0.42\linewidth}
        \centering
        \includegraphics[width=\linewidth]{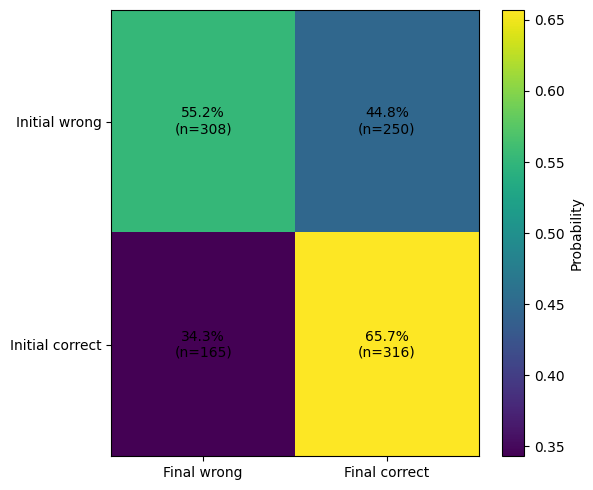}
        \caption{Clean reference graph}
        \label{fig:transition_clean}
    \end{subfigure}
    \hfill
    \begin{subfigure}[b]{0.42\linewidth}
        \centering
        \includegraphics[width=\linewidth]{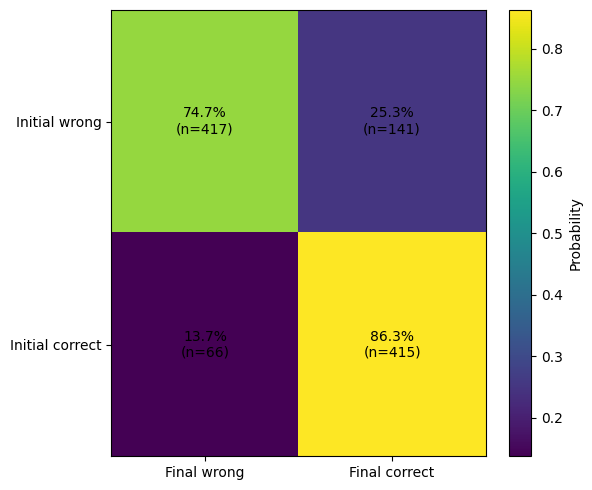}
        \caption{Noisy graph}
        \label{fig:transition_noisy}
    \end{subfigure}

    \caption{Transition matrices between initial embedding-based classification
    and final GSSL classification.
    Rows correspond to initial correctness and columns to final correctness.}
    \label{fig:transition_matrices}
\end{figure}

    \item \textbf{Relation Reconstruction:}
    The relation reconstruction pretext task achieves the best overall performance on the clean graph compared to all other pretext tasks and graph variants. However, its performance degrades significantly on the noisy and refined graph settings, falling below that of the PLM-initialized embeddings. This decline can be attributed to the structural differences between the graphs. The clean graph is organized according to a core ontology, wherein each relation adheres to well-defined domain and range constraints. This structure enables the model to implicitly infer the types of subject and object entities during training. In contrast, the noisy graph constructed using a general-purpose method (GT2KG) lacks ontological structure, exhibits a fragmented and sparse topology, and contains erroneous connections. As a result, the correspondence between relations and entity types deteriorates. Although DistMult remains effective at reconstructing relations under these conditions (Fig.~\ref{fig:relation_reconstruction}), it is not well suited for the typing task due to its limited capacity to model type-specific semantic dependencies.
    \begin{figure}[ht]
    \centering
    \begin{subfigure}[b]{0.48\textwidth}
        \centering
        \includegraphics[width=\linewidth]{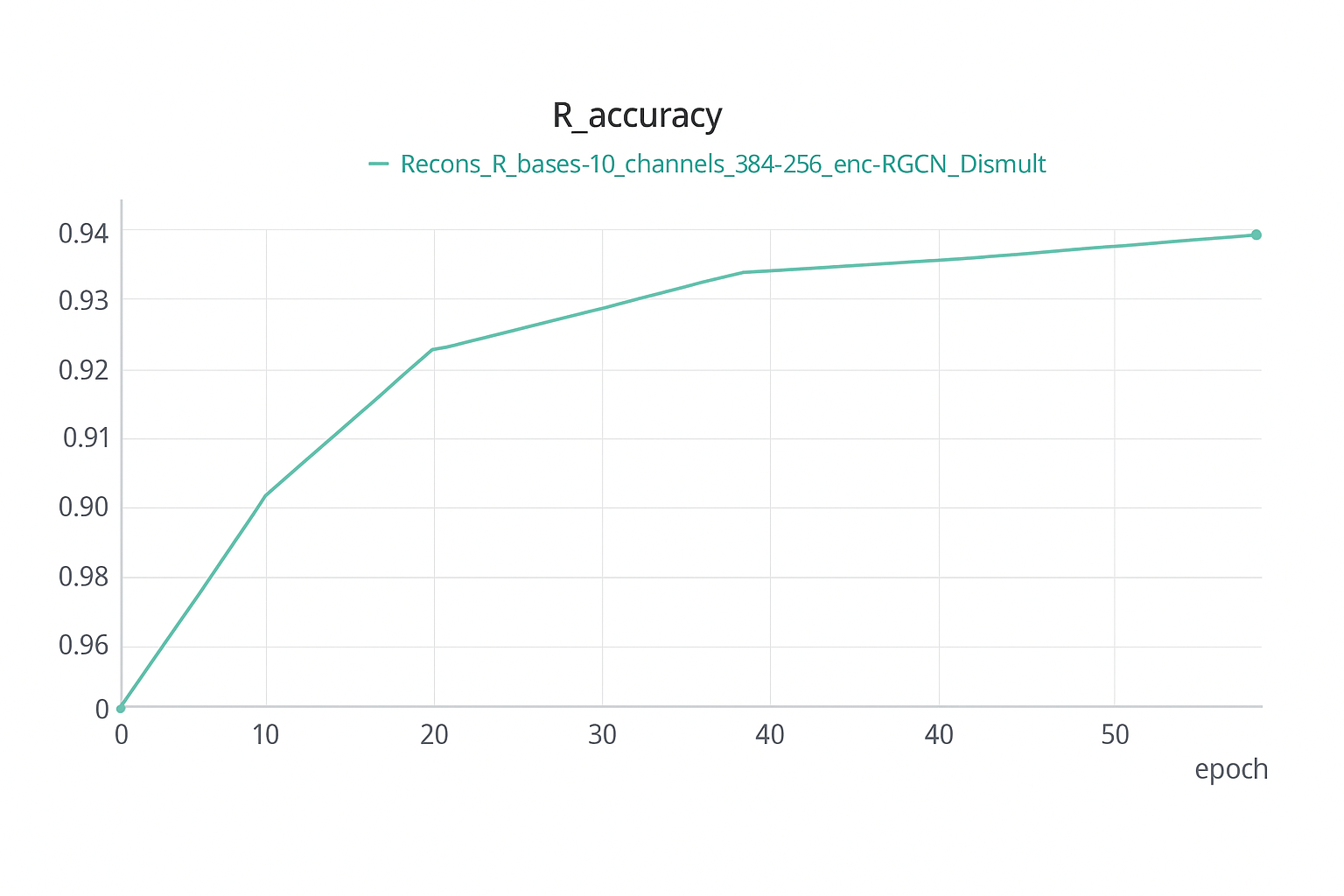}
        \caption{Clean graph}
        \label{fig:clean_recons}
    \end{subfigure}
    \hfill
    \begin{subfigure}[b]{0.48\textwidth}
        \centering
        \includegraphics[width=\linewidth]{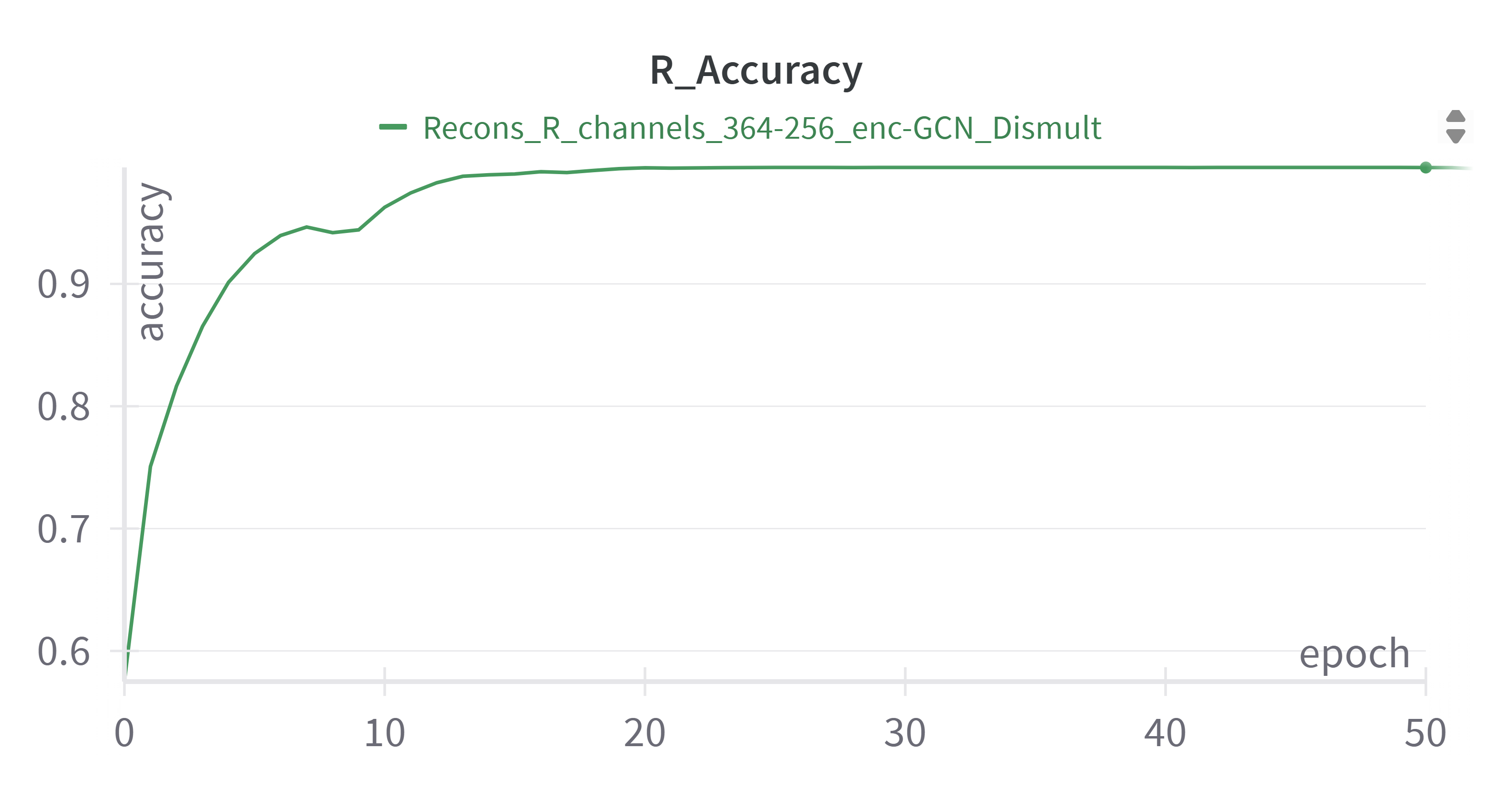}
        \caption{Noisy graph}
        \label{fig:noisy_recons}
    \end{subfigure}
    \caption{Accuracy evolution for the relation reconstruction task
using the DistMult decoder on (a) the clean graph and (b) the noisy graph. }
    \label{fig:relation_reconstruction}
    
\end{figure}
    \item \textbf{Contrastive:}
While contrastive learning is widely recognized for its strong discriminative capacity, our results indicate that it consistently degrades performance relative to the initial embeddings across all graph variants. This uniform drop suggests that the limitation stems not from the quality of the underlying graphs, but from a fundamental misalignment between the contrastive learning objective and the entity typing task.
A key factor contributing to this misalignment is the negative sampling strategy typically employed in contrastive learning methods~\cite{zhu2020deepgraphcontrastiverepresentation}. Specifically, for a given node in one augmented view of the graph, all other nodes in the second view including the predefined type nodes are treated as negatives. As a result, the model is explicitly encouraged to push apart a term and its correct type, which directly contradicts the goal of the typing task. This contradiction leads to degraded performance, as clearly illustrated in Figure~\ref{fig:contrastive_loss_acc}, on the clean graph, the architecture that achieved the highest accuracy corresponds to the one that optimized the contrastive loss the least, further highlighting the incompatibility between the contrastive objective and the typing task.
\begin{figure}[ht]
    \centering
    \includegraphics[width=0.48\textwidth]{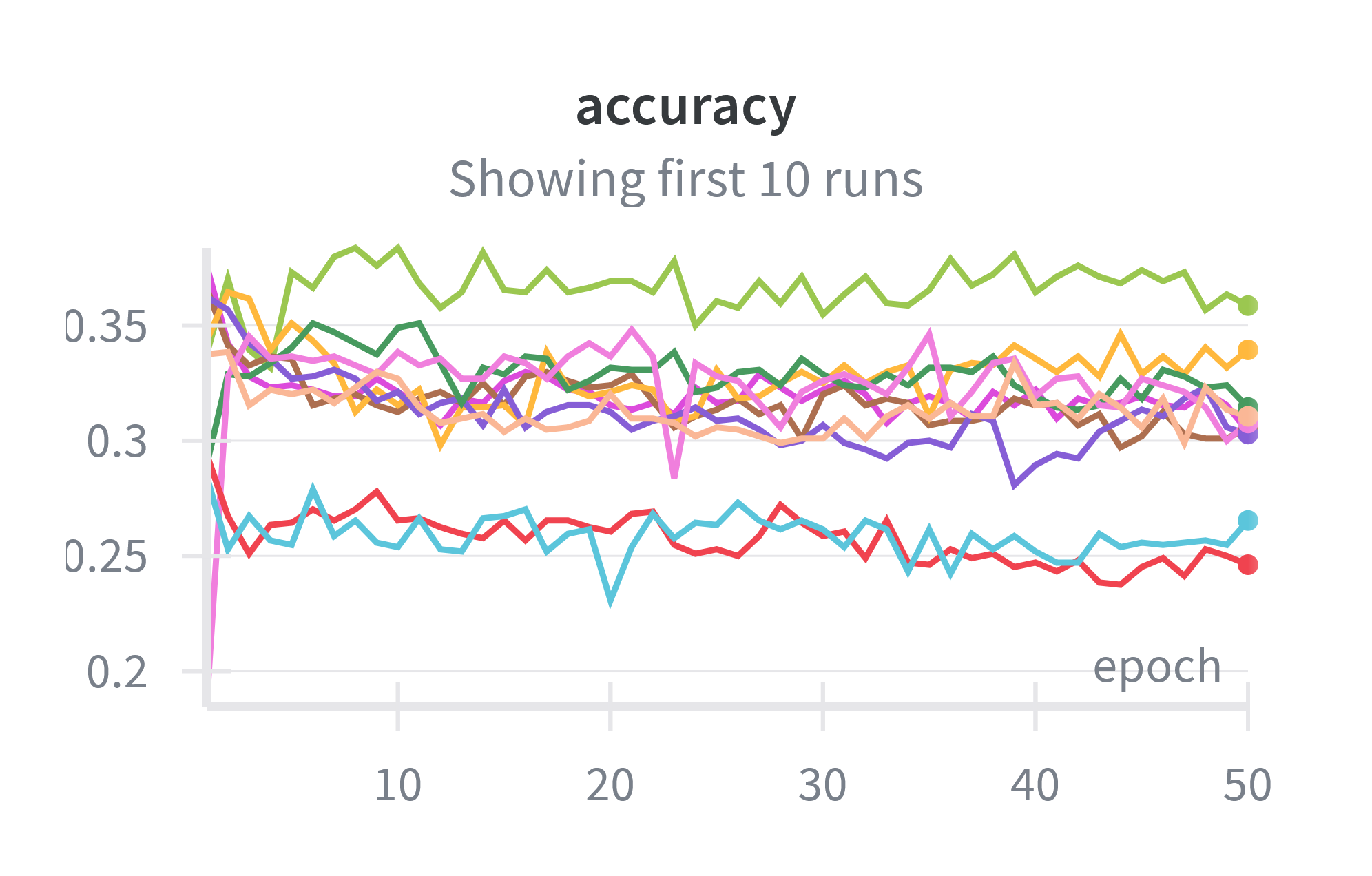}
    \includegraphics[width=0.48\textwidth]{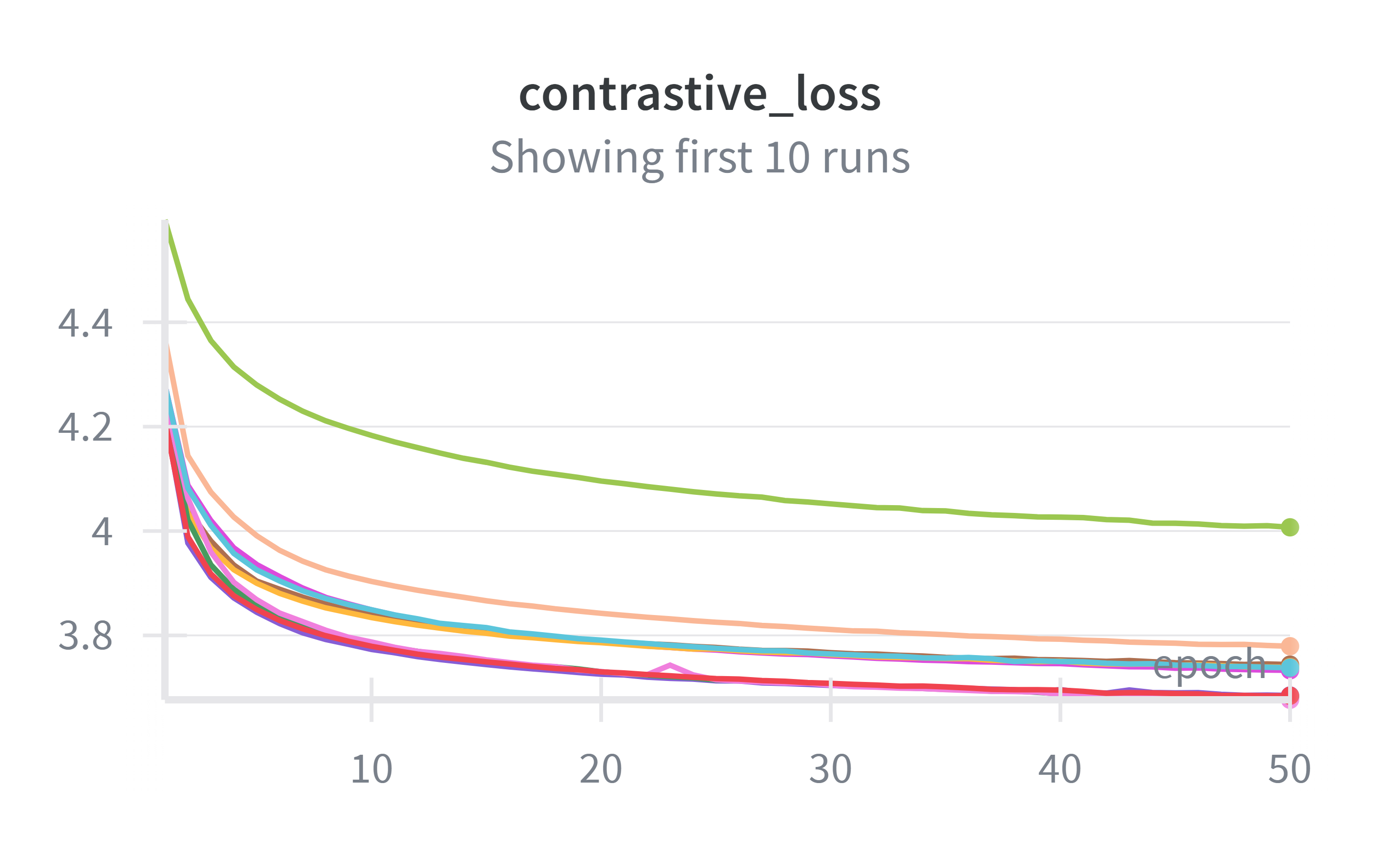}
    \caption{Accuracy (left) and contrastive loss (right) curves for the contrastive learning pretext task across different encoder architectures on the clean reference graph.} 
    \label{fig:contrastive_loss_acc}
\end{figure}

\end{itemize}

\subsubsection{Analysis of Encoder-Decoder Architecture Choices}
\begin{figure}[htbp]
    \centering
    \includegraphics[width=\textwidth]{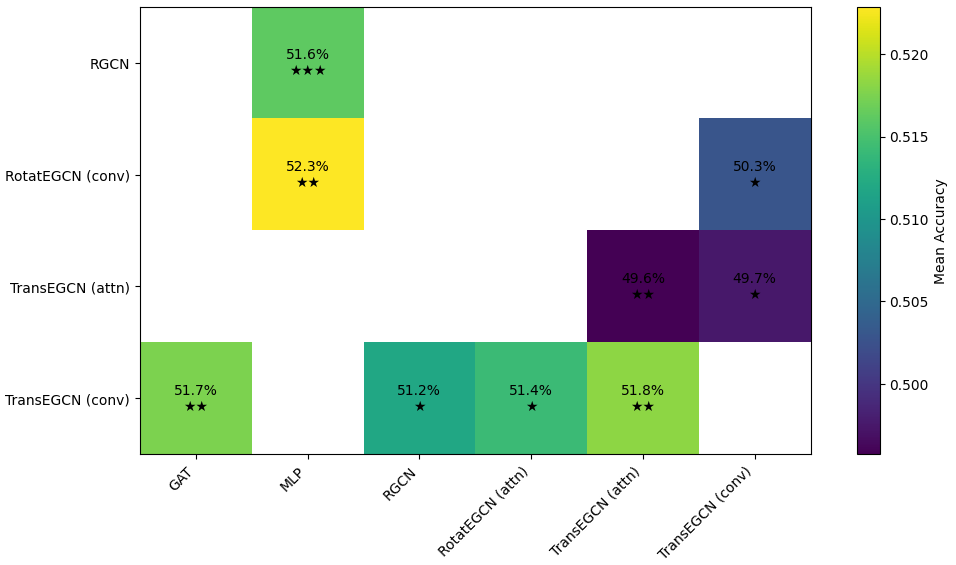}
    \caption{Global heatmap showing encoder-decoder performance across different graph variants. Colors indicate mean accuracy, with ($\star$) indicating the number of graph variants for which the encoder-decoder combination achieves the best performance.}
    \label{fig:global_heatmap_encoder_decoder}
\end{figure}

\begin{figure}[htbp]
    \centering
    \includegraphics[width=\textwidth]{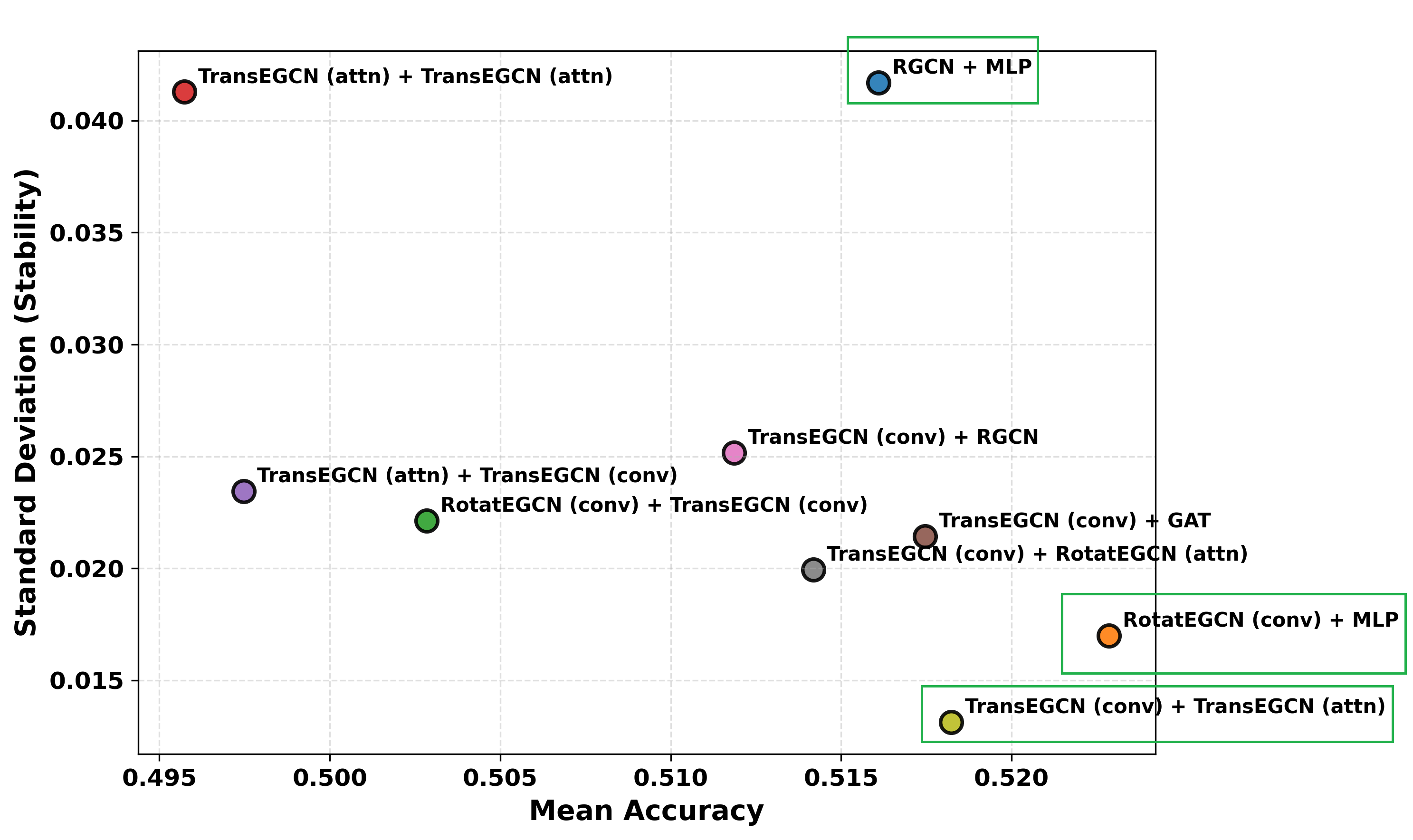}
    \caption{Mean accuracy versus stability (standard deviation) for different encoder-decoder combinations. Highlighted models indicate favorable performance-stability trade-offs.}
    \label{fig:performance_vs_stability}
\end{figure}

We focus here on the architectures associated with the feature reconstruction pretext task, which proved to be the most robust overall.
As shown in Figure~\ref{fig:pretext_accuracy_all}, performance varies depending on the chosen encoder-decoder architecture, and each graph variant exhibits its own set of top-performing combinations.
To analyze how these architectural choices generalize across different graph qualities, we selected the top three performing encoder-decoder combinations for each graph variant. Their performance and stability across all graph variants are summarized in Figure~\ref{fig:global_heatmap_encoder_decoder} and Figure~\ref{fig:performance_vs_stability}.


From these visualizations, three trends emerge:

\begin{itemize}

    \item \textbf{RGCN-MLP}: consistently appears among the top three performers on the \textit{Clean Reference}, \textit{Enriched}, and \textit{Combined Refinement} graphs all of which exhibit high connectivity and low fragmentation. The effectiveness of RGCN-MLP in these settings is attributable to its relational message-passing mechanism, which depends on diverse relation types and aggregates messages from incoming neighbors. However, its performance drops significantly on the \textit{Noisy} and \textit{Cleaned} graphs, which are sparse and structurally degraded. In such graphs, lower in-degree, reduced relational diversity, and isolated components hinder message propagation, negatively impacting contextualization and node typing thereby reducing both accuracy and stability.

    \item \textbf{RotateEGCN(conv)-MLP}: This combination yields the \textit{highest mean accuracy overall} and ranks in the top three for the \textit{Clean Reference} and \textit{Enriched} graphs both well-connected. Notably, despite not ranking in the top three for the \textit{Noisy} and \textit{Cleaned} graphs, it still performs \textit{robustly} in these sparse settings. Its generalization ability stems from the RotateEGCN layers, which integrate \textit{both incoming and outgoing} edges. This bidirectional setting enables even sparsely connected or partially isolated nodes (e.g., those with only one outgoing edge) to receive meaningful information. The architecture therefore demonstrates a favorable trade-off between performance and stability across all graph types.
    
    \item \textbf{TransEGCN(conv)-TransEGCN(attn)}: This architecture is particularly effective for \textit{structurally weak graphs}, ranking among the top three on both the \textit{Noisy} and \textit{Cleaned} variants. Moreover, it performs competitively on augmented or well-structured graphs like the \textit{Clean Reference} and \textit{Enriched} versions. The dual use of TransEGCN layers at both encoder and decoder levels combined with the incorporation of attention mechanisms at the decoder supports robust feature propagation while mitigating over-smoothing in dense graphs. Its ability to leverage both incoming and outgoing edges further enhances its adaptability across different structural conditions.
\end{itemize}

\subsection{Graph Refinement Analysis}
\label{sec:graph_refinement}
\begin{figure}[htbp]
    \centering
    \includegraphics[width=0.75\linewidth]{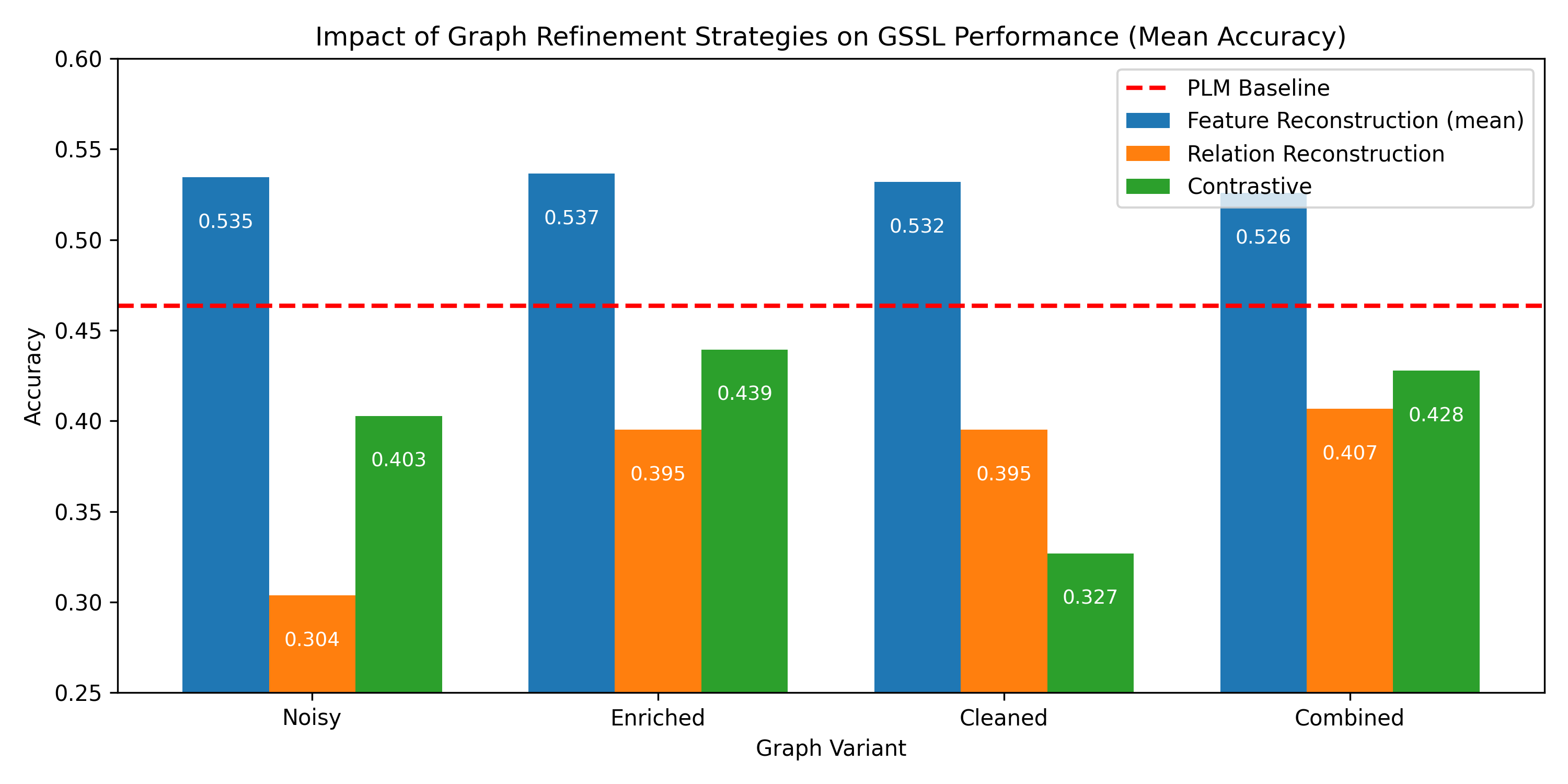}
    \caption{Impact of graph refinement strategies on GSSL performance using mean accuracy per task and graph variant.}
    \label{fig:graph_refinement}
\end{figure}

To evaluate the contribution of graph refinement strategies, we compare the performance of the GSSL framework across four graph variants: \emph{noisy}, \emph{enriched}, \emph{cleaned}, and \emph{combined refined} graphs. Figure~\ref{fig:graph_refinement} summarizes the impact of each refinement strategy on the three pretext tasks: feature reconstruction, relation reconstruction, and contrastive learning.
Graph enrichment substantially improves structural connectivity by reducing graph fragmentation and increasing the giant component ratio. This structural enhancement translates into consistent performance gains across all tasks. Compared to the noisy graph, enrichment improves relation reconstruction accuracy by approximately $+9\%$, contrastive learning by $+3.7\%$, and feature reconstruction by $+0.5\%$. The relatively modest improvement observed for feature reconstruction confirms its robustness to structural noise, whereas relation reconstruction and contrastive learning benefit directly from increased connectivity and more informative neighborhoods.\\
In contrast, graph cleaning exhibits a paradoxical effect. Although it removes noisy or semantically inconsistent edges, it also degrades global connectivity, as reflected by the reduced giant component ratio. This over-conservative edge pruning leads to a severe drop in contrastive learning performance ($-7.6\%$), which strongly depends on neighborhood structure. At the same time, relation reconstruction benefits from the removal of spurious relations, while feature reconstruction remains largely unaffected, further highlighting its limited sensitivity to structural perturbations.\\
The combined refinement strategy yields mixed results. While it achieves the highest accuracy for relation reconstruction (+10\% relative to the noisy graph), its performance for feature reconstruction and contrastive learning is comparable to enrichment alone. This suggests that the additional computational cost of the cleaning step may not be justified.
Overall, these results indicate that for sparse text-driven graphs, structural enrichment appears to be a more reliable refinement strategy than aggressive cleaning.
\section{Conclusion}
Graph Self-Supervised Learning methods offer a wide variety of architectures and learning paradigms capable of deriving meaningful representations from graphs without requiring labeled data. With the growing accessibility of text-driven knowledge graphs, enabled by recent advances in NLP, new opportunities emerge for applying GSSL across multiple domains. However, these opportunities come with the urgent need to understand how GSSL performs on such graphs, which often suffer from complex and heterogeneous noise introduced by automatic construction pipelines, an aspect that most existing robustness studies neglect by focusing only on synthetic noise.
This work presents the first comprehensive study evaluating the performance of various GSSL methods on text-driven graphs for a downstream term typing task. We introduce \textbf{NATD-GSSL}, a unified framework that integrates graph construction, refinement, and GSSL into a single pipeline. To quantify the impact of real-world noise, we propose a \textbf{dual-graph evaluation protocol} that compares GSSL performance on paired noisy and clean graphs, aligned to a shared gold standard within the biomedical domain.
Our results reveal several key findings. First, \textit{relation reconstruction} tasks are highly sensitive to noise and require clean graphs with well-defined schemas, while \textit{feature reconstruction} proves the most robust, achieving performance comparable to that on clean graphs and underscoring the importance of architectural choices. In contrast, contrastive approaches show that robustness depends less on graph quality and more on alignment with the downstream objective. Additionally, our refinement strategies indicate that \textit{graph enrichment} is beneficial for sparse graphs, while over-aggressive \textit{denoising} may degrade performance by reducing structural connectivity. Overall, NATD-GSSL achieves up to \textbf{+7\%} improvement in term typing accuracy compared to pretrained language model baselines.

While our findings provide a solid foundation for understanding GSSL behavior on noisy, text-driven graphs, the study has some limitations. These include the use of a single graph construction method, the challenge of obtaining clean, schema-aligned graphs from text, and refinement techniques evaluated solely through downstream performance. These limitations suggest several promising directions for future research. Future work could extend the evaluation to other domains, downstream tasks, and alternative graph construction pipelines, design contrastive learning strategies with task-aligned negative sampling and contrastive loss formulations, perform fine-grained architectural analyses to separately assess the impact of encoders and decoders, and develop more advanced noise mitigation techniques, including new GNN layers robust to extraction-induced noise. Pursuing these directions with strong empirical grounding will enable the community to move beyond the limitations of prior theoretical work and enhance the practical deployment of GSSL in realistic, noisy, text-driven knowledge graph scenarios.

\section*{Availability of Data and Materials}
All datasets, resources, and source code used in this study are publicly available and fully documented on GitHub at:~\url{https://github.com/OthmaneKabal/MC2GAE}.
The UMLS-NCI graph is not publicly redistributable due to licensing restrictions. It requires an academic research license from the Unified Medical Language System. Therefore, this dataset cannot be shared directly by the authors. However, we provide all scripts and preprocessing pipelines necessary to reconstruct and prepare the graph once the data are obtained from the official source:~\url{https://www.nlm.nih.gov/research/umls/index.html}.

\bibliography{sn-bibliography}

\appendix
\section{PLM Initial Embeddings}
\label{app:plm_init}
This appendix reports the evaluation of initial embeddings obtained from different categories of pre-trained language models on typing task, in order  to identify the most effective representations for initializing graph nodes and relations. 
The results, summarized in Table~\ref{tab:plm_comparison}, provide a comparative overview across general-purpose, domain-specific, and semantic similarity-oriented PLMs.
\begin{table}[ht]
\caption{Evaluation of PLM-based initial embeddings on the term typing task.}
\label{tab:plm_comparison}
{\scriptsize
\begin{tabular}{|c|>{\raggedright\arraybackslash}p{6.2cm}|r|r|r|}
\hline
\textbf{Category} & \textbf{Model} & \textbf{Acc.} & \textbf{F1} & \textbf{Prec.} \\
\hline
\multirow{8}{*}{Semantic Similarity}
& \textbf{all-MiniLM-L6-v2} & \textbf{0.4635} & \textbf{0.4372} & \textbf{0.4731} \\
& paraphrase-MiniLM-L6-v2 & 0.4394 & 0.4211 & 0.4475 \\
& nli-roberta-base-v2 & 0.4163 & 0.3881 & 0.4249 \\
& msmarco-distilbert-base-v3 & 0.3683 & 0.3553 & 0.3587 \\
& multi-qa-MiniLM-L6-cos-v1 & 0.4144 & 0.3905 & 0.4189 \\
& stsb-roberta-base-v2 & 0.3481 & 0.3268 & 0.3438 \\
\hline
\multirow{7}{*}{Domain-specific}
& scibert\_scivocab\_uncased & 0.3981 & 0.3484 & 0.4091 \\
& specter & 0.3404 & 0.3248 & 0.3815 \\
& S-BioBert-snli-multinli-stsb & 0.3913 & 0.3667 & 0.3863 \\
& S-PubMedBert-MS-MARCO & 0.4115 & 0.3539 & 0.4263 \\
& BioLinkBERT-base & 0.1798 & 0.1408 & 0.1908 \\
& biobert-base-cased-v1.1 & 0.2644 & 0.2288 & 0.3776 \\
& Bio\_ClinicalBERT & 0.2538 & 0.2217 & 0.3002 \\
\hline
\multirow{9}{*}{General-purpose}
& bert-base-uncased & 0.2442 & 0.2191 & 0.2952 \\
& bert-base-cased & 0.2712 & 0.2568 & 0.3079 \\
& roberta-base & 0.2144 & 0.1713 & 0.2599 \\
& distilbert-base-uncased & 0.3288 & 0.3069 & 0.3553 \\
& graphcodebert-base & 0.1673 & 0.1483 & 0.1626 \\
& bart-large & 0.2712 & 0.2555 & 0.3149 \\
\hline
\end{tabular}
}
\end{table}

\end{document}